\makeatletter\def\graphicscache@inhibit{true}\makeatother
\documentclass[letterpaper, 10 pt, journal, twoside]{IEEEtran}

\IEEEoverridecommandlockouts

\usepackage{tikz}
\usepackage{tkz-graph}
\usepackage{tikz-3dplot}
\usepackage{pgfplots}
\usetikzlibrary{arrows}
\usetikzlibrary{positioning,calc}
\usetikzlibrary{decorations.pathreplacing}
\usetikzlibrary{decorations.markings}
\usetikzlibrary{fit}
\usetikzlibrary{shapes.callouts}
\usetikzlibrary{shapes.geometric}
\usetikzlibrary{shapes.arrows}
\usetikzlibrary{matrix}
\usetikzlibrary{fit,shapes.callouts,shapes.geometric,backgrounds,positioning,arrows.meta,calc}
\usepackage{pgfplots}
\pgfplotsset{compat=1.9}
\usepgfplotslibrary{groupplots}
\pgfplotsset{every axis/.append style={label style={font=\small},tick label style={font=\small},},}
\usepackage{soul}

\usepackage{algorithm}
\usepackage{algorithmic}
\usepackage[utf8]{inputenc}

\usepackage[hidelinks]{hyperref}
\usepackage[style=ieee,hyperref,natbib=true,backend=bibtex,firstinits,doi=false,%
     mincitenames=1,maxcitenames=2,minbibnames=2,maxbibnames=3,sorting=none,terseinits=false,hyperref=true]{biblatex}
\bibliography{references.bib}
\renewbibmacro*{bbx:savehash}{}%
\defbibheading{bibliography}[\bibname]{\section*{References}}

\usepackage[capitalise]{cleveref}
\usepackage{url}

\usepackage{threeparttable}
\usepackage{booktabs}
\usepackage{multirow}
\usepackage{graphicx}

\newcounter{num}

\graphicspath{{./figures/}}

\usepackage{ifthen}
\IfFileExists{graphicscache.sty}{%
  \usepackage[dpi=440]{graphicscache}%
}{%
}

\def\spacebelowfigures{-0.4cm}
\usepackage{etoolbox}

\title{%
Leveraging Vision-Language Models for Open-Vocabulary\\Instance Segmentation and Tracking
}
\author{Bastian Pätzold$^{*}$$^{1}$$^{2}$$^{3}$, Jan Nogga$^{*}$$^{1}$$^{2}$$^{3}$, and Sven Behnke$^{1}$$^{2}$$^{3}$
\thanks{$^{*}$Equal contribution. Contact: {\tt paetzold@ais.uni-bonn.de}}
\thanks{$^{1}$Autonomous Intelligent Systems, University of Bonn, Germany}
\thanks{$^{2}$Lamarr Institute for Machine Learning and AI, Germany}
\thanks{$^{3}$Center for Robotics, University of Bonn, Germany}
\thanks{Digital Object Identifier: \href{https://doi.org/10.1109/LRA.2025.3606363}{10.1109/LRA.2025.3606363}}
}

\begin{document}

\maketitle
\IEEEpubid{%
    \begin{minipage}{\textwidth}\vspace{1.8cm} \fontsize{6.7}{7.2}\selectfont
        ©~2025 IEEE. Personal use of this material is permitted. Permission from IEEE must be obtained for all other uses, in any current or future media, including reprinting/republishing this material for advertising or promotional purposes, creating new collective works, for resale or redistribution to servers or lists, or reuse of any copyrighted component of this work in other works.
    \end{minipage}
}

\markboth{IEEE Robotics and Automation Letters. November, 2025}{P{\"a}tzold \MakeLowercase{\textit{et al.}}: Leveraging VLMs for Open-Vocabulary Instance Segmentation and Tracking}

\begin{abstract} %

Vision-language models (VLMs) excel in visual understanding but often lack reliable grounding capabilities and actionable inference rates.
Integrating them with open-vocabulary object detection (OVD), instance segmentation, and tracking leverages their strengths while mitigating these drawbacks.
We utilize VLM-generated structured descriptions to identify visible object instances, collect application-relevant attributes, and inform an open-vocabulary detector to extract corresponding bounding boxes that are passed to a video segmentation model providing segmentation masks and tracking.
Once initialized, this model directly extracts segmentation masks, processing image streams in real time with minimal computational overhead.
Tracks can be updated online as needed by generating new structured descriptions and detections.
This combines the descriptive power of VLMs with the grounding capability of OVD and the pixel-level understanding and speed of video segmentation.
Our evaluation across datasets and robotics platforms demonstrates the broad applicability of this approach, showcasing its ability to extract task-specific attributes from non-standard objects in dynamic environments.

\end{abstract}

\begin{IEEEkeywords}
Object Detection, Segmentation and Categorization; Semantic Scene Understanding; Visual Tracking
\end{IEEEkeywords}

\section{Introduction}

\IEEEPARstart{I}{n recent} years, VLMs have emerged as a groundbreaking advancement in computer vision and natural language processing.
Models such as GPT~\cites{gpt4}, Claude~\cites{claude}, Gemini~\cites{gemini}, or Pixtral~\cites{pixtral} have demonstrated remarkable capabilities to comprehend and generate descriptions of complex visual scenes.
One of the key features of these models is their capacity for zero-shot recognition~\cites{zeroshot,zeroshot_vision}, enabling them to perceive a wide range of visual scenes without prior training on a particular problem domain.
Moreover, they exhibit a unique ability to provide rich, contextually aware interpretations that are readily directed to extract task-relevant information and adhere to structural requirements.

While a naive VLM-generated image description may, to some extent, be considered sufficient for solving various perception tasks, we identify three core issues that must be addressed for application in a robotics context.
First, the time required to generate such a description exceeds the typical inference time of a traditional vision pipeline.
This directly limits their applicability to time critical tasks in dynamic environments.
Second, the description is unstructured natural language intended for human readers.
Consequently, the relevant object information cannot be directly parsed into a machine-readable format for downstream use.
Third, the description cannot be readily associated with the corresponding image content, lacking the grounding capabilities provided by traditional vision pipelines.
However, dense pixel-level understanding is often required for downstream interaction or manipulation tasks.

In this work, we address these issues to unlock the visual perception capabilities of VLMs for robotics applications.
We achieve this by integrating VLMs with other vision modules to describe and recognize object instances.
Our method provides task-relevant attributes, bounding boxes, and segmentation masks at high frame rates, without requiring training or fine-tuning for specific scenarios or object sets, relying solely on widely available general-purpose foundation models.

\begin{figure}[t!]
    \centering
    \includegraphics[width=1.0\linewidth]{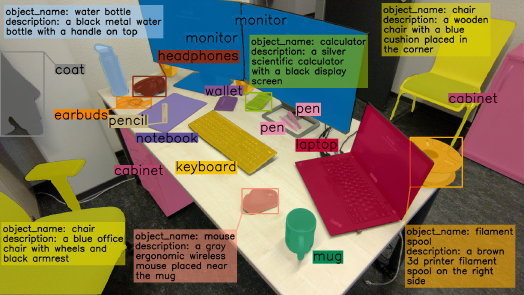}
    \vspace{-0.6cm}
    \caption{Visualization of detected object instances, including object classes, visual descriptions, bounding boxes, and segmentation masks, using off-the-shelf foundation models without prior knowledge of the image content. Note, that some object descriptions are collapsed to reduce visual clutter.}
    \vspace{\spacebelowfigures}
    \label{fig:teaser}
\end{figure}

Our contributions include:
\begin{enumerate}
    \item a method integrating off-the-shelf models to robustly identify, describe, ground and track objects in a scene,
    \item an instance-aware assignment scheme curating the output of any given open-vocabulary detector,
    \item an evaluation protocol for grounded object descriptions compatible with existing object detection benchmarks,
    \item a comparison with baselines in real-world experiments on a mobile manipulator, on a custom dataset featuring diverse non-standard objects, and on a public dataset.
\end{enumerate}
For full reproducibility of results, we release \textit{VLM Grounding for Instance Segmentation \& Tracking (VLM-GIST)}\footnote{\label{fn:vlm}\url{https://vlm-gist.github.io}}, including our implementation, evaluation scripts and custom dataset.

\section{Related Work}

\subsection{Closed-Vocabulary Detection in Robotics}

Robots require robust perception capabilities to effectively interact with their environment, as understanding and interpreting the surrounding space is prerequisite to executing tasks.
A key aspect of robotic perception is object (and person) detection, which is crucial for manipulation and interaction tasks.
Traditional detection methods often assume that the objects of interest belong to a predefined set known during training.
While effective, these closed-vocabulary detectors face significant limitations: they struggle to recognize unseen instances of a known class and are unable to detect objects belonging to entirely unknown classes.
Moreover, they primarily assign class labels to bounding boxes and fail to extract richer semantic information.
Popular datasets like COCO~\cites{coco} and specialized benchmarks like the YCB Object and Model Set~\cites{ycb} have driven advancements in these detectors but are inherently limited by their fixed taxonomies.

\subsection{Open-Vocabulary Detectors}

The introduction of aligned image and text encodings such as CLIP~\cites{clip} has made room for object detectors generalizing beyond a fixed set of classes.
Since then, performance in open-vocabulary detection has scaled with training schemes which increasingly exploit image-level data annotation.
In this vein, Detic~\cites{zhou2022detectingtwentythousandclasses} combines object detection samples with region proposals and class labels from image classification data.
OWLv2~\cites{minderer2023scalingopenvocabularyobjectdetection} consumes image captions by grounding their N-grams to obtain pseudo-annotated samples for object detection.
To allow for training on phrase grounding and referring expression comprehension datasets, Grounding DINO~\cites{liu2023groundingdino} utilizes an attention mask based on phrase extraction to generate sub-sentence text encodings,
while OmDet-Turbo~\cites{zhao2024realtimetransformerbasedopenvocabularydetection} reformulates samples from object detection, image caption, human-object interaction and phrase grounding datasets as visual question answering tasks to train on.
More recently, LLMDet~\cites{fu2025llmdet} co-trains with a VLM to benefit from detailed image and region level annotations.

While suitable for constrained robotics applications~\cites{winner_paper}, these models tend to overemphasize individual nouns, neglect relational terms~\cite{zhao2024mmgroundingdino}, and struggle with commonsense knowledge such as logos~\cites{detector_eval}.
In practice, this necessitates careful manual prompt design around the expected objects and their attributes, a limiting factor in open robotics environments.

\subsection{Instance Segmentation and Tracking}

Object instance masks for RGB-D data yield finer object point clouds, which are useful for downstream tasks such as grasp pose estimation.
These can be obtained by using bounding boxes to prompt Segment Anything (SAM)~\cites{kirillov2023segment}, an image segmentation foundation model, as demonstrated by Grounded SAM~\cites{ren2024groundedsam}.
In the context of agentic systems, the ability to track object masks over time is desirable to close the feedback loop between object interactions.
This is readily achieved in a robotics use case with RGB-D sensors based on object class and robot localization, but requires frequently running detector inference.
This overhead can be avoided by instead prompting SAM 2~\cites{ravi2024sam2}, a recent video segmentation foundation model which runs on a mobile GPU.

\subsection{Manual-Free Segmentation}

Recent methods relax requirements for manual prompting.
RAM-Grounded-SAM~\cites{ren2024groundedsam} grounds image tags without accounting for specific object attributes.
GenSAM~\cites{hu2023GenSAM} obtains image-specific object prompts, localized to use as SAM prompts. An iterative reweighting scheme then refines object heatmaps.
ProMaC~\cites{hu2024ProMaC} generates patch captions, initializing an iterative reasoning process using inpainting to prune hallucinations.
These methods rely on CLIP for semantic disambiguation and cannot yield fully structured instance segmentations of cluttered environments.

\subsection{Vision-Language Models}

Models such as Kosmos-2~\cites{peng2023kosmos2} or Florence-2~\cites{xiao2023florence2} offer captioning into phrase grounding and partially address the issues of open-vocabulary detectors.
However, trained on a discrete set of task templates, they lack the ability to extract specific semantic information in a machine-readable format.

In contrast, integrating vision capabilities on top of state-of-the-art Large Language Models (LLMs), i.e. VLMs~\cites{gpt4,claude,gemini,pixtral}, leverages the world knowledge and perception capabilities encoded in the LLM and allows for nuanced interactions~\cites{visionllm}.
In addition to visual question answering, this enables constraining the response format or engaging in multi-turn conversations.
\citet{molmo} propose Molmo, a generalist family of VLMs with a special focus on visual grounding, capable of pointing to 2D pixel locations supporting a given answer.
Similarly, \citet{qwen} propose Qwen-VL with an emphasis on bounding box detection, introducing special tokens that allow association with corresponding descriptive text fragments.
We believe that these models are currently bottlenecked by processing visual tokens projected from image patches rather than dense pixel-wise representations, resulting in inaccurate bounding boxes.

While we find such extensions to the core formulation of a VLM intriguing, we focus on generalist models that provide a standard set of features and are easily interchangeable.
Our motivation is to take advantage of the rapid advances in available models while reducing system complexity and memory requirements, since a single model can be used for other tasks beyond instance segmentation and tracking.

\section{Method}

\subsection{Overview \& Application}

\begin{figure}[t!]
    \centering
    \includegraphics[width=1.0\linewidth]{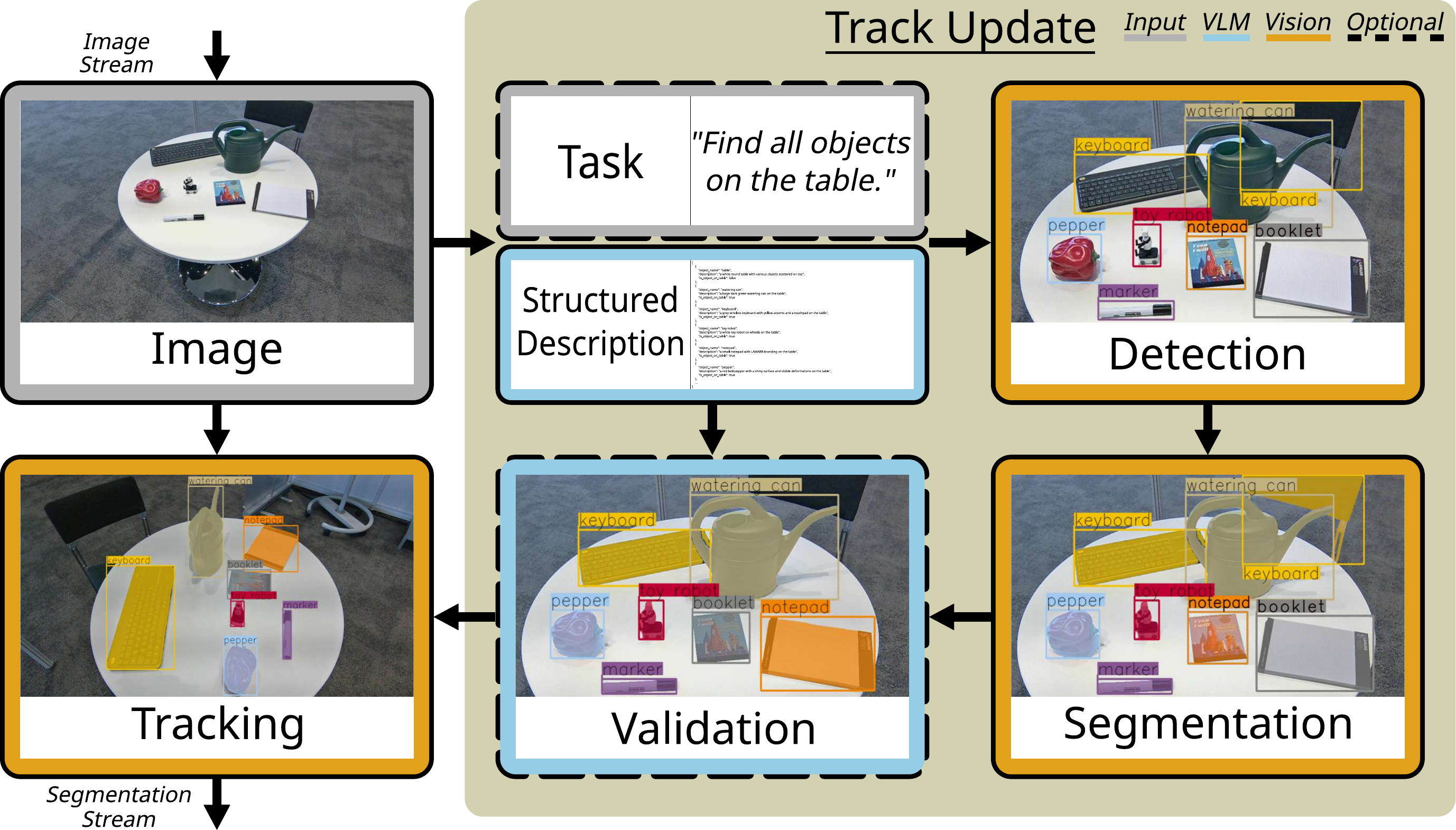}
    \vspace{-0.6cm}
    \caption{Pipeline leveraging vision-language models for open-vocabulary instance segmentation and tracking. The low-frequency update mechanism initializes the tracker and updates tracks on demand. The lightweight tracker generates high-frequency segmentation masks on the full image stream.}
    \vspace{\spacebelowfigures}
    \label{fig:pipeline}
\end{figure}

\cref{fig:pipeline} illustrates our method.
It comprises an update mechanism and a direct path between an incoming image stream and a tracker.
To start continuously tracking object instance masks with corresponding semantic information, the tracker is initialized by passing a single image through the update mechanism.
First, the update mechanism uses a VLM to generate a \textit{structured description} -- a machine-readable list that provides semantic attributes and a natural language description of all object instances visible in the image.
Then, the natural language descriptions are used to ground the object instances through an open-vocabulary detector.
Finally, the resulting bounding boxes are passed on to a segmentation model to obtain the corresponding masks and update the tracker.
Since the update mechanism is comparatively slow, it is not intended to be applied on every incoming frame, but whenever new object instances become visible.
It can be executed on demand within the context of the application or at regular intervals.
The tracker, on the other hand, is lightweight and capable of processing an entire image stream, given common frame rates and resolutions.

\subsection{Structured Description}

To effectively use an open-vocabulary detector, careful prompt design is required to explicitly ground all objects in an image.
This requires incorporating scene knowledge, including object characteristics and relationships.
However, this dependency on prior knowledge limits its applicability in open robotics environments.
Our goal is to automate this process, approximating an oracle that provides suitable detector prompts without requiring prior scene knowledge.

\subsubsection{Model Selection}

We use a VLM to generate a structured description.
All of our VLM interactions generate text from text and image input.
To extract detector prompts for all visible object instances, we require the VLMs' responses to be machine-readable.
To accomplish this, we employ the established method of requesting the VLM to respond with JSON-compliant text~\cites{json_1,json_2}.

\subsubsection{Prompting Technique}

We prompt a VLM to generate a single list in JSON format, with a separate entry for each unique object visible in the image.
Each entry must be a dictionary containing multiple key-value pairs, i.e. \textit{attributes}, that provide necessary information both to the user and to subsequent pipeline steps.
In particular, we require the attributes \textit{object\_name} and \textit{description}, where the first is used like a class label in a typical detector and the second is used to prompt an open-vocabulary detector.
Its ability to recognize an object instance is directly affected by specifying the expected format and content of the \textit{description} attribute.
\citet{detector_eval} compare the sensitivity of several open-vocabulary detectors to different types of prompts.
They show how some detectors benefit from including spatial relationships to other image features in the prompt, while others benefit from including color or material.
We explicitly prompt the model to provide unique instance descriptions, limited to ten words, that include type, color, and appearance.

Additional attributes can be defined by the user to extract features that are relevant in the context of the application.
They can be configured to require certain attributes, restrict the validity of values and data types, or establish dependencies between attributes.
Adapting them before each execution of the update mechanism allows to dynamically shift the attention of the system's visual perception.
The time required to generate a structured description is directly proportional to its length, the type of model, and the hardware used.
While it's length depends on the number of object instances visible in the image, limiting the number and conciseness of attributes is key to minimizing update-latency.
\cref{fig:description} shows several examples of object instances contained in a structured description with four additional user-defined attributes.

\subsubsection{Post-Processing}

We apply various techniques to robustly obtain a valid structured description from a raw VLM response.
First, we use regular expressions to extract the longest valid JSON object in order to accommodate artifacts such as additional text or Markdown tags outside of the JSON body.
Then, we process each list element that is a dictionary and discard all others.
We verify that the expected attributes exist and that their values are valid.
To account for minor inaccuracies and minimize unnecessary filtering, we employ further correction techniques, which include typecasting and expected keyword matching based on the Levenshtein distance.
Finally, we ensure uniqueness of \textit{object\_name} attributes by appending sequential numbers to duplicates.

\subsubsection{Decoupled Attribution}

\begin{figure}[t!]
    \centering
    \includegraphics[width=1.0\linewidth]{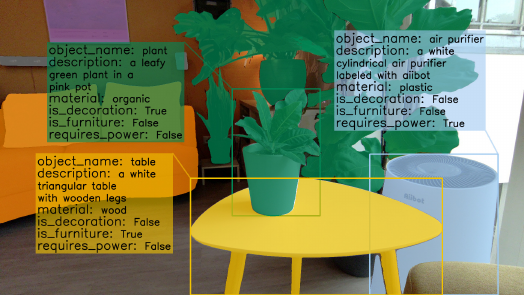}
    \vspace{-0.6cm}
    \caption{Object instances with four additional user-defined attributes parsed from a structured description annotating their corresponding detections.}
    \vspace{\spacebelowfigures}
    \label{fig:description}
\end{figure}

The described approach to attribution works well for simple attributes like those shown in \cref{fig:description}.
However, we observe that the structured description tends to be strongly biased by prompts that contain more complex attribution instructions.
For example, when we specify the Boolean attribute \textit{task\_relevant} for the task ``\textit{Find a pen}'', we observe that the considered VLMs tend to identify similar or hardly visible objects as pens, though they describe them accurately in the absence of this attribute.

To address this issue, we allow the user to define a concise natural language task and add the Boolean attribute \textit{task\_relevant} to each object instance.
Its generation is decoupled from the initial structured description in a subsequent prompt.
In particular, we prompt the model to generate a JSON-encoded list of all task-relevant object instances, referenced by their \textit{object\_name} attribute.
Using the same post-processing techniques mentioned before, we augment the initial structured description with the additional attribute.
Note that this approach addresses bias while also reducing the number of output tokens and enabling parallel attribution.

\subsection{Open-Vocabulary Detection}

Once a structured description containing all the desirable semantic information for all visible object instances in the image is obtained, we proceed to ground each one.
First, the \textit{description} attributes from all object instances are collected and used as prompts for an open-vocabulary detector.
In principle, any text-promptable detector that can produce bounding boxes is suitable for this task.
We use MM-Grounding-DINO~\cites{zhao2024mmgroundingdino}, which is a strong open-weight model trained on an open mixture of datasets.

Since the detector can recall more than one object per prompt, while our goal is to obtain a one-to-one mapping to object instances in the structured description, superfluous detections must be discarded.
The subsequent validation step is also designed to achieve this regardless of the number of detections it is fed.
We balance the ratio of detections that are immediately discarded and those that are discarded downstream by defining the \textit{Over-Detect-Factor} (ODF).
This factor scales the number of object instances in the structured description to provide an upper bound on the number of detections propagated.
In particular, we initially select the most confident detection for each object instance.
For an ODF\,$>$\,1, we add detections until the upper bound is reached, prioritizing confident ones.
In this case, skipping the optional validation step will result in duplicate detections, which increases recall and reduces precision.

\subsection{Instance Segmentation \& Tracking}

We use SAM 2~\cites{ravi2024sam2} to obtain segmentation masks for all object instances based on the bounding boxes provided by the detector.
We then track the object instances across an image stream using its video segmentation capability.

Since the official SAM 2 implementation does not support causal inference, we utilize a third-party repository~\cites{sam2_realtime}.
If the tracker is not initialized, we prompt it with all bounding boxes from the detection step.
This yields track IDs and segmentation masks corresponding to each detection.
On the other hand, when established tracks exist, we retrieve the current tracker state, convert all masks to bounding boxes and suppress new track creation for detections with an IoU $>0.6$ relative to an existing track.
We then concatenate the filtered new boxes with the previous ones to update the tracker state.

\subsection{Validation \& Error Correction}

The optional validation step aims to increase confidence in the mapping between descriptions and object masks.
It does this by either validating, correcting, or rejecting the previously obtained grounding of all object instances, while ensuring that each object instance is grounded at most once.
This is useful when precision is preferred over recall, for example, if a robot is supposed to interact with any one of multiple objects, it is unnecessary to ground each one, but beneficial to prioritize correctly grounded objects.

\subsubsection{Generate Validation Labels}

For each grounded object instance, we prompt a VLM with the full image, a crop of the full image at the respective bounding box, the full structured description, and an instruction to respond with the \textit{object\_name} attribute of the object instance seen in the crop.
If the crop cannot be associated with any of the object instances, we instruct the model to respond with a special \textit{invalid} keyword instead.
If the response is not \textit{invalid}, it yields a new proposed mapping between a description and an object track.
All object instances can be processed in parallel, so the effective validation latency is the maximum response time for any one proposal.

\subsubsection{Solve Assignment Problem}

To decide whether to validate, correct, or reject a proposal, we collect the responses for all object instances and compare them to the original mapping.
We group object instances with similar \textit{object\_name} attributes, so that each object instance can be uniquely associated with a group, and each group has one or more object instances associated with it.
This helps to disambiguate object instances that are difficult to visually distinguish based on their crops.
Finally, we use a heuristic to maximize the agreement between the validation responses and the original mapping.
This is achieved, for example, by resolving duplicate agreements between both mappings to unassigned group members, using the detector confidence of the original mapping, and adhering to the validation response when prior steps are inconclusive.

\section{Implementation}

For full reproducibility and to facilitate further research and benchmarks, we publish \textit{VLM-GIST}\textsuperscript{\ref{fn:vlm}} including our implementation, evaluation scripts and custom dataset.

\subsubsection{VLM Inference \& Middleware}

For local inference, we orchestrate multiple vLLM servers with a custom load balancing module, which routes API requests and scales the maximum processable requests with the number of nodes.

To interface with a wide range of VLMs, we developed a ROS2 package that acts as a middleware for interfacing with the Chat Completions API established by OpenAI.
It supports commercial online APIs such as OpenAI, Mistral AI, and OpenRouter, as well as the open source inference framework vLLM~\cites{vLLM}, which allows us to self-host (quantized) open-weight models from families such as Pixtral~\cites{pixtral}, Molmo~\cites{molmo}, Qwen~\cites{qwen}, and InternVL~\cites{internvl}.
By handling aspects such as monitoring, timeout behavior, model parameters, error correction, and asynchronous generation, this package makes it easy to run, switch, and compare models in a general, robust, reproducible, and consistent manner.
We also use this package to interface all other models used in the evaluation, which can either run locally on a robot or on a remote server.

\begin{figure}[t!]
    \centering
    \begin{tikzpicture}
        \node[anchor=south west, inner sep=0pt] (image1) at (0,0) {\includegraphics[width=0.713\linewidth]{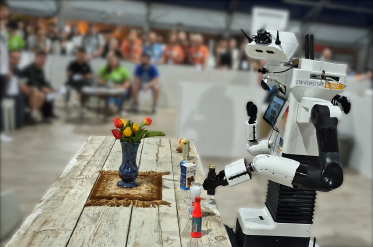}};
        \node[anchor=south west, inner sep=1pt, fill=white, opacity=0.8, text opacity=1] at (image1.south west) {\small a)};
    \end{tikzpicture}
    \hfill
    \hspace{-0.5cm}
    \begin{tikzpicture}
        \node[anchor=south west, inner sep=0pt] (image2) at (0,0) {\includegraphics[width=0.2826\linewidth]{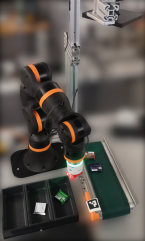}};
        \node[anchor=south west, inner sep=1pt, fill=white, opacity=0.8, text opacity=1] at (image2.south west) {\small b)};
    \end{tikzpicture}
    \vspace{-0.2cm}
    \caption{Experimental robot platforms. (a) Team NimbRo's TIAGo++ robot at the RoboCup@Home 2024 finals in Eindhoven, grasping ingredients associated with a dinner recipe. (b) Industrial scenario where objects on a conveyor belt are detected from above, picked up, and placed in boxes.}
    \vspace{\spacebelowfigures}
    \label{fig:robots}
\end{figure}

\subsubsection{Domestic Service Robot}

We developed a modified PAL Robotics TIAGo++ omnidirectional two-armed platform (see \cref{fig:robots}a) to compete in the RoboCup@Home Open Platform League 2024~\cites{winner_paper}.
The purpose of this competition is to promote the development of general-purpose autonomous service robots that can be deployed in challenging and unknown domestic environments.
It is equipped with an Orbbec Gemini 2 RGB-D camera running at 1080p.
For VLM inference, we rely on online APIs accessed through an onboard 5G router.
The detector and tracker run locally alongside several other compute-intensive tasks on a Zotac ZBOX with an NVIDIA RTX A4500 Mobile 16GB GPU.

\subsubsection{Industrial Grasping Robot}

In collaboration with igus~GmbH, the Lamarr Institute, Fraunhofer IAIS and IML, we developed a robot demonstrator (see \cref{fig:robots}b) that mimics an industrial scenario.
In particular, an igus ReBeL 6-DoF arm equipped with a suction gripper is placed next to a conveyor belt to pick up items from it and place them in nearby storage containers.
A generic 4K webcam is used to detect the objects on the conveyor belt from a top-down perspective.

\section{Evaluation}

\subsection{Robot Experiments}

\subsubsection{Qualitative Experiments}

Our method was used in the winning performance of Team NimbRo in the finals of the RoboCup@Home Open Platform League 2024~\cites{winner_paper}.
In this 10-minute demonstration, our robot explored an arena simulating a typical apartment and helped to prepare dinner.

First, the robot explored the apartment and collected images from potential food locations.
The images were passed to the update mechanism and processed using \textit{GPT-4o-2024-05-13} while exploration continued.
We defined the task \textit{"Find all food and cooking ingredients."} to identify all available food items and discard all other objects, such as furniture and decorations.
The robot then received a dinner order from one of the judges, for which we generated a recipe based on the available ingredients in the apartment.

Then the robot moved to one of the food locations and executed the update mechanism with the new task \textit{"Grasp the ingredients mentioned in the following recipe: ..."}.
By evaluating the \textit{task\_relevant} attributes of the detected object instances to identify the food items contained in the recipe, the robot tried to grasp two of them using both of its arms.
Between this initial detection and each grasp, the robot had to adjust its relative position to the objects several times to accommodate the arm's workspace and to find a collision-free grasping trajectory.
The tracking capability of our method is essential here because it eliminates the need to detect and associate target objects reliably several times in a row.
Finally, the robot could deliver the ingredients and inform the judge of their nature and purpose in the recipe by evaluating the \textit{description} attributes of the grasped objects.

In the industrial scenario, we developed an LLM-based agent that attempts to execute any given natural language instruction using a set of pre-defined tools for perception (our method), grasping, placing, and logging.
The goal was to demonstrate how a robot in a typical industrial scenario can use and benefit from modern approaches for object recognition and natural language understanding.
Tasks this system performed include sorting chocolates by color, finding the food item with the oldest expiration date, and classifying products by arbitrary properties.
For example, it can reason that the chocolate packaged in white is likely not vegan because its label suggests that it contains yogurt, which is usually made from milk.
This shows how our method can handle uncommon scenarios in terms of the top-down view, the industrial setting, and a variety of unknown products.
It also demonstrates the ability to read text, recognize brands and logos, and interpret images on packaging materials.

\begin{table}[t!]
    \caption{Robot Evaluation Results.}
    \label{tab:robot_experiments}
    \centering
    \setlength{\tabcolsep}{2pt}
    \begin{threeparttable}
        \begin{tabular}{l|c|c|c||c|c|c|c}
            \toprule
            \multicolumn{2}{c|}{Experiments} & %
            \% Task &
            \% Obj. &
            \multicolumn{4}{c}{\% of Failures} \\
            \cmidrule{5-8}
            \multicolumn{2}{c|}{} &
            Succ.\,$\uparrow$ &
            Ident.\,$\uparrow$ &
            Agent & Nav. & Manip. & Vision \\
            \midrule
            \multirow{2}{*}{Baseline} & Ours     & 85 & 90 &  0 & 33 & 67 &  0 \\
                                      & Kosmos-2 & 35 & 25 &  8 &  8 & 15 & 69 \\
            \midrule
            \multirow{1}{*}{Complex} & Ours     & 67 & 87 & 40 &  0 & 20 & 40 \\
            \bottomrule
        \end{tabular}
        \begin{tablenotes}\footnotesize
            \item Performance for a mobile manipulator performing a simple baseline task and more complex tasks. Performance is measured by successful task execution and percentage of correctly identified target objects. Task failures cases are assigned to the responsible robot sub-component, where vision refers to our pipeline.
        \end{tablenotes}
        \vspace{\spacebelowfigures}
    \end{threeparttable}
\end{table}

\subsubsection{Quantitative Experiments}

\cref{tab:robot_experiments} quantifies the performance of our method on a real robot.
For the experiments, we deploy our method as the vision system for a mobile manipulator with an agentic task planner.
To account for the performance of other necessary systems, we design a simple baseline task: \textit{"Bring me the object on the couch\,\slash\,pantry table."}.
The object is not specified, and there is only a single object at the specified location, so the vision system must only ground and track (for realignment in manipulation) the object, but does not necessarily need to identify it accurately.
Under these requirements, we do not expect our method to fail, which places the full burden on the planning, navigation and manipulation stacks.
We select one location with a low grasping surface and one with a high one to cover the robot's reachability.
Ten household items which fit in the robot's gripper are used as test objects, covering various shapes and sizes.
We perform this task for each combination of object and location for a total of 20 trials.
A trial is counted as a success if it ends with the robot successfully handing the object to the operator.
For failed trials, we assign the cause to the responsible sub-component.
Additionally, the object is considered correctly identified if its description is accurate.

\begin{figure}[t!]
    \centering
    \includegraphics[width=1.0\linewidth]{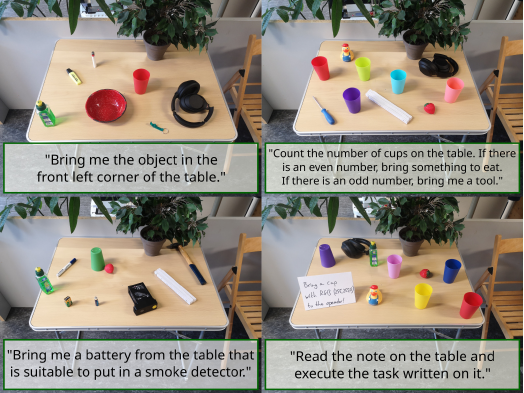}
    \vspace{-0.6cm}
    \caption{Manipulation scenes and corresponding tasks which were successfully solved by a mobile manipulator equipped with our proposed method.}
    \vspace{\spacebelowfigures}
    \label{fig:pantry}
\end{figure}

We observe that failures result only from navigation or manipulation errors, but never due to our vision system.
The objects are always grounded and mostly correctly identified, e.g. in one instance the robot brought a benchy boat but described it as an origami figurine.

We repeat the experiment with Kosmos-2~\cites{peng2023kosmos2} replacing the VLM and open-vocabulary detector.
The task success rate is low, with most failures originating from the vision system.
Very small objects are missed, so the robot cannot interact with them.
When an object is grounded, the bounding boxes are often imprecise, leading to a bad tracker initialization and subsequent tracking failures.
Even when the object is grounded and retrieved, it is most often described incorrectly.
Overall, despite the model's performance in other tasks, we find it is not suitable for this application.
We also attempted to use Florence-2~\cites{xiao2023florence2}, but found that it described singular objects on furniture along with the furniture itself, grounding both together, such that manipulation is not possible.

Finally, we explore the limitations of the proposed method by applying it to 15 more complex tasks (see \cref{fig:pantry}) including semantic disambiguation (e.g. \textit{bringing specifically the upside down cup among several cups}),
logic and counting (e.g. \textit{bringing a certain object if there is an odd number of objects with a specific attribute on the table and another object otherwise}),
and reading comprehension (e.g. \textit{reading a note at the location and executing the task described on it}).
We also use a larger set of objects and introduce clutter unrelated to the task.
All tasks are formulated such that an object must be retrieved for the operator, which ensures that the grounding and tracking components of our method are tested as well.
Here, due to the complexity of the tasks, planning is more likely to fail when objects are too vaguely described by our method, e.g. the agent is searching for cereals but only a 'white box' is described.
Our method also fails several times when the open-vocabulary detector confuses visually similar objects, e.g. swaps the detection of a power cable with a nearby HDMI cable, or when tracks are lost on small objects in a cluttered scene.
Overall, observing a success rate of $67\%$ compared to an ideal result of $85\%$ given the performance of other sub-components, we find that our method can address a wide variety of challenging tasks.

\subsection{Vision Benchmarks}

\subsubsection{Label Matching}

Since our method assigns open-set names and descriptions with each detection, there is no direct way to evaluate performance metrics on public datasets.
To achieve this, we need to develop a mechanism to automatically find the semantic associations between our detections and the annotated classes contained in the dataset.
We do this by first asking \textit{GPT-4o-2024-11-20} to generate five sentence definitions of the abstract category referred to by each of our detections and the annotated classes in the dataset.
For the former, we provide the LLM with the corresponding \textit{object\_name} and \textit{description} attributes from the structured description.
We then use OpenAI's \textit{text-embedding-3-large} model to encode both attributes and definition of all detections, and the class name and definition of all dataset classes.
Finally, for all detections and classes, we compute the pairwise cosine similarities between their embeddings, average them to obtain a single similarity score for each pair, and find the most similar class for each detection.

\begin{figure}[t!] %
    \centering
    \begin{tikzpicture}
        \node[anchor=south west, inner sep=0pt] (image1) at (0,0) {\includegraphics[width=0.495\linewidth]{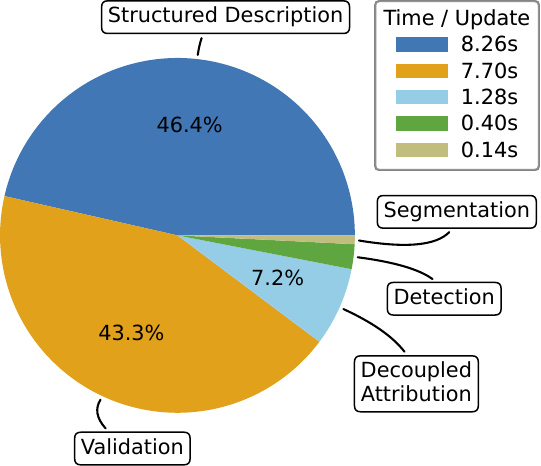}};
        \node[anchor=south west, inner sep=1pt, fill=white, opacity=0.8, text opacity=1] at (image1.south west) {\small a)};
    \end{tikzpicture}
    \hfill
    \hspace{-0.2cm}
    \begin{tikzpicture}
        \node[anchor=south west, inner sep=0pt] (image2) at (0,0) {\includegraphics[width=0.495\linewidth]{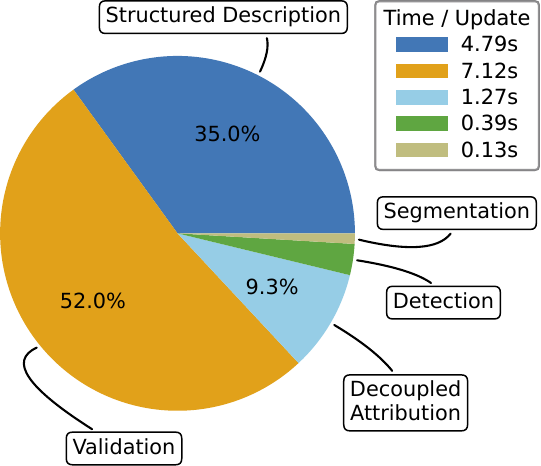}};
        \node[anchor=south west, inner sep=1pt, fill=white, opacity=0.8, text opacity=1] at (image2.south west) {\small b)};
    \end{tikzpicture}
    \vspace{-0.555cm}
    \caption{Partial times for execution of the update mechanism. Experimental setup: Using (a) \textit{GPT-4o-2024-11-20} and (b) \textit{Gemini 2.0 Flash} (description \& validation) and \textit{MM-GDINO-L zero-shot} with an Over-Detect-Factor of 1.5 and one decoupled attribute on our custom dataset. Each structured description contains (a) $\mu=13.8$ and (b) $\mu=12.3$ instances.}
    \vspace{\spacebelowfigures}
    \label{fig:time}
\end{figure}

The class set of a dataset is most often incomplete, i.e. it contains object instances that are not annotated because they do not correspond to any class in the scope of the dataset.
Therefore, we must decide to either assign each detection to the most similar class or discard it from further evaluation.
Instead of thresholding the obtained similarities, we augment the set of classes with additional ones that are designed to have minimal semantic overlap with the dataset classes, reducing the  incompleteness.
Then, if the most similar class to a detection is an augmented one, we discard the detection from further evaluation.
Otherwise, we assign the respective native dataset class and evaluate it accordingly.

Finally, we introduce mapping rules to address systematic annotation problems where a category of depicted objects is repeatedly annotated with a dissimilar class.
For example, soap dispensers might be annotated as \textit{bottle}, rather than not being annotated due to a lack of an appropriate dataset class.
Here we augment the missing object categories as described, but map matched detections to the respective dissimilar dataset classes instead.
This scheme, parametrized by augmented labels, definitions and their corresponding text embeddings, wraps detection metrics to enable evaluation of manual prompting-free methods on object detection benchmarks.

All reported performance metrics are obtained using this approach.
For COCO minival, we use 271 augmented classes and eleven mapping rules.
For our densely annotated custom dataset with unique descriptions instead of classes, we do not use augmented classes, but match against all descriptions contained in the image.
In a manual verification of 100 random object instances described on COCO images, we find 89\% correct matches and 11\% borderline cases where the instance description is not sufficient to uniquely assign it to one of the COCO classes.

\begin{table}[t!]
    \caption{Isolated Detector Metrics.}
    \label{tab:dataset}
    \centering
    \setlength{\tabcolsep}{2.44pt}
    \begin{threeparttable}
        \begin{tabular}{l|c|ccc|cc|cc|ccc}
            \toprule
                \multirow{2}{*}{\raisebox{-0.5\height}{Detector}} &
                \multirow{2}{*}{\raisebox{-0.5\height}{Data}} &
                \multicolumn{3}{c}{mAP} &
                \multicolumn{2}{c}{Precision} &
                \multicolumn{2}{c}{Recall} &
                \multicolumn{3}{c}{F\textsubscript{1}-score} \\
            \cmidrule(lr){3-5} \cmidrule(lr){6-7} \cmidrule(lr){8-9} \cmidrule(lr){10-12}
                                    &        &       All &      Cls  &       Ins &       Cls &       Ins &      Cls  &      Ins &       All &      Cls  &      Ins \\
            \midrule
            \multirow{2}{*}{MMGD}   & Custom &     n.a.  & \bf{0.36} & \bf{0.43} & \bf{0.55} & \bf{0.54} & \bf{0.46} & \bf{0.54} &     n.a.  & \bf{0.50} & \bf{0.54} \\ %
                                    & COCO   & \bf{0.50} &     0.51  & \bf{0.56} & \bf{0.81} & \bf{0.78} &     0.70  & \bf{0.78} & \bf{0.73} &     0.75  & \bf{0.78} \\ %
            \midrule
            \multirow{2}{*}{LLMDet} & Custom &     n.a.  &     0.33  &     0.41  &     0.45  &     0.52  &     0.45  &     0.52  &     n.a.  &     0.45  &     0.52  \\ %
                                    & COCO   &     0.49  & \bf{0.53} & \bf{0.56} & \bf{0.81} & \bf{0.78} & \bf{0.72} & \bf{0.78} &     0.71  & \bf{0.76} & \bf{0.78} \\ %
            \midrule
            \multirow{2}{*}{OmDetT} & Custom &     n.a.  &     0.16  &     0.20  &     0.16  &     0.28  &     0.28  &     0.28  &     n.a.  &     0.20  &     0.28  \\ %
                                    & COCO   &     0.33  &     0.38  &     0.42  &     0.71  &     0.65  &     0.54  &     0.65  &     0.57  &     0.61  &     0.65  \\ %
            \midrule
            \multirow{2}{*}{OWLv2}  & Custom &     n.a.  &     0.08  &     0.09  &     0.17  &     0.14  &     0.12  &     0.14  &     n.a.  &     0.14  &     0.14  \\ %
                                    & COCO   &     0.38  &     0.41  &     0.45  &     0.46  &     0.72  &     0.39  &     0.72  &     0.41  &     0.42  &     0.72  \\ %
            \bottomrule
        \end{tabular}
        \begin{tablenotes}
            \item Performance metrics for isolated open-vocabulary detectors. We prompt detectors with \textbf{all} classes, those in the image (\textbf{cls}), and the corresponding \textbf{ins}tance counts. Conf. thresh. for \textit{All} and \textit{Cls} maximize the F\textsubscript{1}-score.
        \end{tablenotes}
        \vspace{\spacebelowfigures}
    \end{threeparttable}
\end{table}

\subsubsection{Datasets}

Our custom dataset contains diverse, non-standard objects in domestic environments that resemble challenging manipulation scenes.
It contains 64 images, partially drawn from AgiBot World~\cite{agibot}, with an average of 18 annotations (range 8 to 48), covering 136\% (cumulative) of each image.
We exhaustively annotated all visible objects in each image, assigning natural language descriptions that address their type, appearance, and location -- sufficient for unique identification.
COCO minival contains 5000 images with an average of 7 annotations, covering 54\% (cumulative) of each image.
Of these images, 1\% have no annotations, 54\% have up to five, and 15\% have 15 or more.

\subsubsection{Experiments}

To assess the practicality of our method, we examine the execution time of the update mechanism.
\cref{fig:time} shows that it is dominated by (partially optional) VLM interactions, while detection and segmentation are negligible.
While we find the use of larger models with the optional steps feasible in practice, a small quantized model can generate structured descriptions in 2.5s, even for complex scenes. %

In \cref{tab:dataset} we report common performance metrics of several open-vocabulary detectors in isolation.
We evaluate each detector with increasing degrees of information about a given image.
Specifically, we first evaluate under the established protocol of prompting with all class labels occurring in the dataset.
Since the labels in our custom dataset represent instance-specific descriptions instead of class labels, this is not applicable for our dataset.
Next, we apply a class label oracle providing the ground truth labels occurring for each image.
In either case, we report all metrics for the detector-specific confidence threshold which maximizes the F\textsubscript{1}-score for a given dataset.
Finally, we apply an instance-aware oracle that additionally provides the number of occurrences per label, wrapping each detector with an ODF\,=\,1.0.

For all detectors, we observe a significant performance drop when evaluating on our custom dataset.
This is expected, as it contains more complex scenes and more fine-grained annotation.
Overall, MM-Grounding-DINO performs best, and we thus use it for all subsequent experiments.
It is worth noting that LLMDet does not convincingly outperform MM-Grounding-DINO, despite co-training with a VLM.
It may be limited by its architecture (same as MM-Grounding-DINO) or its training scale.

\begin{table}[t!]
    \caption{Baseline Comparison.} %
    \label{tab:baselines}
    \centering
    \setlength{\tabcolsep}{4.6pt}
    \begin{threeparttable}
        \begin{tabular}{c|l|c|c||c|c|c|c}
            \toprule
                \multirow{2}{*}{\raisebox{1.5\height}{Data}} &
                \multirow{2}{*}{\raisebox{1.5\height}{Model}} &
                \multicolumn{1}{c| }{\scalebox{1.0}[1.0]{Ins.}} &
                \multicolumn{1}{c||}{\scalebox{1.0}[1.0]{Time}} &
                \multicolumn{1}{c| }{\scalebox{1.0}[1.0]{mAP}} &
                \multicolumn{1}{c| }{\scalebox{1.0}[1.0]{Pre.}} &
                \multicolumn{1}{c| }{\scalebox{1.0}[1.0]{Rec.}} &
                \multicolumn{1}{c  }{\scalebox{1.0}[1.0]{F\textsubscript{1}}} \\
            \midrule

            \multirow{16}{*}[-0.0pt]{\rotatebox[origin=c]{90}{COCO}}
            & Gemini 2.5 Pro               &     11.1  &     13.9  & \bf{0.35} &     0.60  &     0.49  & \bf{0.54} \\
            & Gemini 2.5 Flash             &     9.24  &     3.48  &     0.32  &     0.59  &     0.44  &     0.51  \\
            & Gemini 2.0 Flash             &     6.00  &     3.41  &     0.30  &     0.49  &     0.44  &     0.46  \\
            & GPT-4.1                      &     10.5  &     8.72  &     0.33  &     0.55  &     0.47  &     0.51  \\
            & GPT-4.1 mini                 &     7.81  &     5.22  &     0.30  &     0.62  &     0.43  &     0.50  \\
            & GPT-4o                       &     8.83  &     4.96  &     0.30  &     0.49  &     0.43  &     0.46  \\
            & OVIS 2.5 9B                  &      7.02  &      6.97  &     0.32  &     \bf{0.67} &     0.40  &     0.50  \\ %
            & Mistral Medium 3.1           &      8.40  &      3.21  &     0.31  &     0.61  &     0.41  &     0.49  \\ %
            & Qwen2.5-VL-72B\dag           &      5.88  &     13.3  &     0.33  &     0.54  &     0.46  &     0.49  \\

            & \scalebox{0.9}[1.0]{InternVL2.5-38B-MPO\dag}      &     7.26  &     16.2  &     0.33  &     0.49  &     0.49  &     0.49  \\
            & \scalebox{0.9}[1.0]{InternVL2.5-8B-MPO\dag\ddag}  &     5.12  &     2.32  &     0.26  &     0.54  &     0.37  &     0.44  \\
            & \scalebox{0.9}[1.0]{InternVL2.5-4B-MPO\dag\ddag}  &     5.54  &     1.80  &     0.28  &     0.50  &     0.36  &     0.42  \\
            & \scalebox{0.9}[1.0]{InternVL2.5-1B-MPO\dag}       &     3.55  & \bf{0.81} &     0.22  &     0.50  &     0.25  &     0.33  \\
            & Claude Sonnet 4              &     7.93  &     5.64  &     0.29  &    0.65 &     0.37  &     0.47  \\
            & Grok-2                       &     7.85  &     4.83  &     0.30  &     0.43  &     0.48  &     0.45  \\
            \cmidrule(lr){2-8}
            & Florence-2                   & \bf{11.2} &     0.95  &     n.a.  &     0.49  & \bf{0.55} &     0.52  \\ %
            & Kosmos-2                     &     4.58  &     1.60  &     n.a.  &     0.44  &     0.19  &     0.26  \\

            \midrule

            \multirow{6}{*}[-0.0pt]{\rotatebox[origin=c]{90}{Custom}}
            & Gemini 2.5 Pro               & \bf{14.3} &     15.7  & \bf{0.36} &     0.52  & \bf{0.40} & \bf{0.45} \\
            & Gemini 2.5 Flash             &     12.9  &     4.37  &     0.30  &     0.49  &     0.34  &     0.40  \\
            & GPT-4.1                      &     13.8  &     10.7  &     0.31  &     0.48  &     0.36  &     0.41  \\
            & GPT-4.1 mini                 &     10.3  &     6.78  &     0.28  & \bf{0.57} &     0.32  &     0.41  \\
            & OVIS 2.5 9B                  &      8.59  &      8.51  &     0.24  & \bf{0.57} &     0.27  &     0.36  \\ %

            \cmidrule(lr){2-8}
            & Florence-2                   &     6.97  & \bf{0.82} &     n.a.  &     0.33  &     0.13  &     0.18  \\ %
            & Kosmos-2                     &     4.63  &     1.27  &     n.a.  &     0.26  &     0.07  &     0.11  \\

            \bottomrule
        \end{tabular}
        \begin{tablenotes}
            \item Grounded \textbf{ins}tances with inference time, and performance metrics for description models and baselines. We use ODF\,=\,1.0 without validation on 500 COCO minival images and our custom dataset. Highlighted models are (\dag) inferred locally and (\ddag) 4-bit AWQ-quantized~\cites{awq}.
        \end{tablenotes}
        \vspace{\spacebelowfigures}
    \end{threeparttable}
\end{table}

On both datasets, all detectors benefit from per-image prompting represented by the class and instance-aware oracles.
Intuitively, using more information about the image leads to better scores, but this result also confirms that our proposed assignment scheme using the ODF exploits the label counts while absolving practitioners of tuning detector confidence thresholds.
Of course, the instance-aware oracle is not available in practice.
We argue that this is not a drawback because in open robotics environments, manually crafting suitable object prompts beforehand is unrealistic.
Thus, parsing a given scene is a necessary step in any case, which justifies the proposed approach to approximate the instance-aware oracle using a VLM.

\cref{tab:baselines} compares our method to Kosmos-2~\cites{peng2023kosmos2} and Florence-2~\cites{xiao2023florence2}.
We present results for COCO minival and our custom dataset, as well as for several VLMs used to generate the structured descriptions within our method.
In line with general VLM benchmarks~\cites{chatbot_arena}, we observe that the performance of our method improves with larger or more recent models.
While the performance of Florence-2 is similar to the best models used in our method for COCO minival, the performance of both baselines drops significantly on our custom dataset.
This is due to the lack of rich semantic content in the descriptions they generate, which are insufficient for associating them with semantically rich and fine-grained ground truth descriptions.
Instead, our method's ability to maintain performance under this condition confirms its practical feasibility in open robotics environments.
Note that we use the same prompts for all VLMs in all experiments, allowing for further optimization.

Additionally, we measure the effect of the validation step for multiple models and observe the intended precision -- recall tradeoff.
For example, for GPT-4o on COCO minival, recall decreases from $0.43$ to $0.36$ while precision increases from $0.49$ to $0.64$.
Thus, users are advised to apply the optional validation based on their priorities.

\section{Conclusion}

Our proposed method \textit{VLM-GIST} leverages foundation models to generate structured descriptions from images that identify all visible object instances and extract arbitrary user-defined attributes, obtain corresponding bounding boxes and segmentation masks, and efficiently track them in an image stream.
We show that our approach generalizes across model variants and that its performance is directly correlated with the general capabilities of the models used.
We demonstrate how \textit{VLM-GIST} can be effectively integrated into real-world robotics applications, suitable for use in conjunction with LLM-based agents.
Moreover, it is capable of automatically annotating datasets without human involvement.
Such datasets may prove useful for improving the performance of open-vocabulary detectors or VLMs in this or other applications.

\section*{Acknowledgement}

We thank the anonymous reviewers for their valuable feedback.
This research has been partially funded by the Federal Ministry of
Education and Research of Germany under grant no. 01IS22094A WestAI.

\printbibliography

\end{document}